\documentclass[conference]{IEEEtran}

\usepackage{booktabs}
\usepackage{multirow}
\usepackage{graphicx}
\usepackage{listings}
\usepackage{tikz}
\usetikzlibrary{positioning, shapes.geometric, arrows.meta, shadows,
                decorations.pathmorphing, calc, fit, backgrounds}
\usepackage{algorithm}
\usepackage{algpseudocode}
\usepackage{amsmath}
\usepackage{amssymb}
\usepackage[skip=4pt]{caption}
\usepackage{xcolor}
\usepackage[table]{xcolor}
\usepackage{url}
\usepackage{hyperref}
\usepackage[most]{tcolorbox}
\usepackage{cite}

\hypersetup{
  colorlinks=true,
  linkcolor=blue,
  urlcolor=blue,
  citecolor=blue
}

\definecolor{vibrantblue}{RGB}{70,130,200}
\definecolor{softbluebg}{RGB}{230,242,255}

\usetikzlibrary{positioning, shapes.geometric, arrows.meta, shadows,
                decorations.pathmorphing, calc, fit, backgrounds}

\lstdefinelanguage{Prolog}{
  morekeywords={:-, use_module, library, integer, is},
  sensitive=true,
  morecomment=[l]{\%},
}
\tcbset{
  mybox/.style={
    colback=blue!5,
    colframe=blue!40,
    boxsep=1pt,
    left=2pt, right=2pt, top=2pt, bottom=2pt,
    width=\linewidth,
    sharp corners,
    enhanced,
    fonttitle=\bfseries,
    breakable
  }
}
\newtcolorbox[auto counter, number within=section]{myexample}[2][]{%
  mybox, title={#2}, #1
}

\begin{document}

\title{NeuroProlog: Multi-Task Fine-Tuning for Neurosymbolic Mathematical Reasoning via the \textit{Cocktail} Effect}

\author{\IEEEauthorblockN{Pratibha Zunjare}
\IEEEauthorblockA{\textit{Electrical and Computer Engineering} \\
\textit{Virginia Tech}\\
Blacksburg, USA\\
pratibhaz@vt.edu}
\and
\IEEEauthorblockN{Michael S. Hsiao}
\IEEEauthorblockA{\textit{Electrical and Computer Engineering} \\
\textit{Virginia Tech}\\
Blacksburg, USA \\
hsiao@vt.edu}
}

\maketitle

\begin{abstract}
We present \textbf{NeuroProlog}, a neurosymbolic framework that compiles math word problems into executable Prolog programs and trains LLMs to internalize symbolic structure rather than apply it post-hoc. Our multi-task \textit{Cocktail} objective jointly supervises formula-to-rule translation~(KB) and natural language-to-program synthesis~(SOLVE) within a shared symbolic space, combining declarative knowledge with procedural reasoning supervision improves compositional reasoning. At inference, an execution-guided decoding pipeline with a five-class error taxonomy enables zero-shot iterative program repair without correction-specific training.

On GSM8K across four model scales (3B--32B), \textit{Cocktail} FT yields statistically significant gains for three of four models: +5.23\% for Qwen-32B, +3.43\% for GPT-OSS-20B, and +5.54\% for Llama-3B; GPT-OSS-20B Cocktail FT achieves the highest accuracy among the evaluated configurations at 88.34\%. Error analysis reveals scale-associated differences in failure and repair behavior across the evaluated models. These results demonstrate the potential of combining symbolic knowledge and execution feedback to improve LLM-based mathematical reasoning. 

\end{abstract}


\section{Introduction}

Large Language Models (LLMs) have achieved strong performance on mathematical reasoning benchmarks such as GSM8K~\cite{gsm8k}, aided by techniques like Chain-of-Thought prompting~\cite{wei2022cot} and Program-of-Thoughts~\cite{poT2023}. However, recent studies reveal that LLM reasoning often relies on probabilistic pattern matching rather than formal logical inference~\cite{lyu2023faithful, valmeekam2022plan}. This leads to brittle behavior: models produce plausible but incorrect solutions, fail under perturbations, and cannot verify intermediate reasoning steps~\cite{creswell2022selection}.

Neurosymbolic approaches attempt to address these limitations by integrating neural models with symbolic representations or external solvers, combining neural flexibility with symbolic rigor~\cite{creswell2022selection, singh2026verge, chen2025neurosymbolic}. However, existing methods typically apply symbolic reasoning as a \textit{post-hoc correction mechanism} at inference time---validating or refining LLM outputs using external theorem provers or logic engines~\cite{logic-lm, linc}. This decoupled design prevents models from internalizing symbolic structure during training: the symbolic components serve as verification tools rather than learned reasoning capabilities. Consequently, LLMs remain weak at systematic generalization to novel problem compositions and struggle to produce verifiable, executable reasoning traces without external intervention.

\begin{figure}[t]
\centering
\begin{tikzpicture}[
    scale=0.5,
    transform shape,
    box/.style={rectangle, draw, rounded corners, minimum width=3cm, minimum height=1.2cm, align=center, font=\footnotesize\bfseries, thick},
    task/.style={box, fill=blue!15},
    data/.style={box, fill=green!15 },
    model/.style={box, fill=orange!15},
    arrow/.style={->, thick}
]

\node[font=\scriptsize\bfseries] at (-3.5,4.2) {Traditional Approach};
\node[data] (trad-data) at (-3.5,3) {Problem-Solving\\Examples Only};
\node[model] (trad-model) at (-3.5,1.2) {LLM};
\node[task] (trad-verify) at (-3.5,-0.6) {External Solver\\(post-hoc)};

\draw[arrow] (trad-data) -- (trad-model);
\draw[arrow] (trad-model) -- node[right, font=\tiny] {verify} (trad-verify);
\draw[arrow, dashed, red] (trad-verify.east) to[out=0,in=0]
    node[right, font=\scriptsize, text=red] {no learning}
    (trad-model.east);

\node[font=\footnotesize\bfseries] at (3.5,4.2) {\textbf{NeuroProlog} (Ours)};
\node[data] (kb-data) at (1.2,3) {Knowledge\\Base};
\node[data] (solve-data) at (5.8,3) {Problem-\\Solving};
\node[model] (our-model) at (3.5,1.2) {LLM $f_\theta$};
\node[task] (exec) at (3.5,-0.6) {Prolog\\Executor $\sigma$};

\draw[arrow, blue!60] (kb-data.south) --
    node[pos=0.5, left=2pt, font=\scriptsize, align=center]
    {symbolic\\grounding}
    (our-model.north west);

\draw[arrow, blue!60] (solve-data.south) --
    node[pos=0.5, right=2pt, font=\scriptsize, align=center]
    {procedural\\demos}
    (our-model.north east);

\draw[arrow] (our-model) -- (exec);

\draw[arrow, dashed, green!60!black] (exec.west) to[out=180,in=180]
    node[left=1pt, font=\scriptsize, text=green!60!black, align=center]
    {error\\feedback}
    (our-model.west);
\end{tikzpicture}
\caption{Comparison of traditional LLM fine-tuning with \textit{NeuroProlog}.}
\label{fig:overview}
\end{figure}
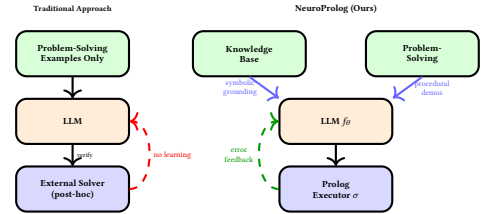

Grounded mathematical reasoning, we posit, requires models to jointly learn three core capabilities: (i) mapping natural language to formal logic, (ii) generating executable programs that encode structured reasoning, and (iii) aligning symbolic outputs with numeric verification. To address this gap, we present \textbf{NeuroProlog}, a unified neurosymbolic framework that enforces formal reasoning through multi-task \emph{Cocktail} training~\cite{brief2024mixing} as shown in Fig.~\ref{fig:overview}. Our approach supervises LLMs on formula translation, program synthesis, and execution verification simultaneously within a shared symbolic representation space, inducing cross-task transfer and enabling models to internalize systematic reasoning patterns rather than relying on superficial heuristics.

We construct a multi-task dataset combining symbolic mathematical knowledge and word problems with executable program traces. Symbolic structure paired with semantic alignment guides the model to generate interpretable, verifiable outputs. At inference, an execution-guided decoding pipeline provides iterative feedback to correct program errors, measuring not only final accuracy but also the model's ability to recover from logical mistakes. The main contributions of this work are as follows:

\begin{itemize}
\sloppy
    \item \textbf{Multi-Task Neurosymbolic Training}: A \textit{Cocktail} objective combining declarative knowledge (formula-to-Prolog) with procedural problem-solving (word-problem-to-program), inducing cross-task transfer within a unified symbolic representation space.

    \item \textbf{Execution-Guided Decoding}: An iterative refinement pipeline with a 5-class error taxonomy enabling zero-shot execution-guided program repair without correction-specific training.

    \item \textbf{Scale-Dependent Error Shift}: Empirical discovery that \textit{Cocktail} training qualitatively transforms error distributions at large scale (TYPE\_ERROR $\rightarrow$ DOMAIN\_ERROR at 32B) but not at small scale (SYNTAX $\rightarrow$ TYPE at 8B), revealing a capacity threshold ($\sim$10B parameters) for semantic type understanding.

    \item \textbf{Comprehensive Evaluation}: Rigorous experiments across 12 configurations (4 models $\times$ 3 settings) demonstrating statistically significant gains (+3.43\% to +5.54\%, $p{<}0.05$) for 3 of 4 models, with strongest performance (88.34\%) exceeding all comparable baselines.

    \item \textbf{Open Release}: Complete dataset (200 KB entries + 310 SOLVE problems + 7476 updated GSM8K-Prolog problems), training code, evaluation pipeline, and LoRA adapters for all 8 fine-tuned configurations.
\end{itemize}

\section{Related Work}
\label{sec:related}

\noindent\textbf{Mathematical Reasoning in LLMs.}
Recent LLMs achieve strong performance on mathematical benchmarks through intermediate reasoning techniques. Chain-of-Thought (CoT) prompting~\cite{wei2022cot} generates natural language rationales with extensions including zero-shot CoT~\cite{kojima2022large} and self-consistency decoding~\cite{wang2023selfconsistency}. Program-Aided Language Models (PAL)~\cite{gao2023pal} and Program-of-Thoughts (PoT)~\cite{poT2023} delegate computation to Python interpreters, generating executable code as reasoning traces. However, these approaches operate at inference time through prompting rather than internalizing symbolic reasoning during training. Studies show LLMs perform probabilistic pattern matching rather than systematic reasoning~\cite{lyu2023faithful,valmeekam2022plan}, failing under distribution shifts~\cite{razeghi2022impact}. We address this through symbolic supervision during fine-tuning, teaching models to generate verifiable logical programs.

\noindent\textbf{Neurosymbolic AI.}
Neurosymbolic approaches combine neural learning with symbolic logic~\cite{garcez2019neuralsymbolic,kautz2022third}. Logic-LM~\cite{logic-lm} uses external Prolog solvers, Faithful CoT~\cite{lyu2023faithfulcot} translates reasoning to formal logic for verification, and LINC~\cite{linc} learns neural-symbolic translations through self-supervised refinement. ProSLM~\cite{vakharia2024proslm} integrates Prolog with language models for explainable, domain-specific question answering. These methods apply symbolic reasoning post-hoc at inference time. Prolog's declarative semantics suit mathematical reasoning, and GSM8K-Prolog provides Prolog annotations for GSM8K with single-task supervision. We extend this through multi-task \textit{Cocktail} training with semantic alignment, internalizing symbolic structure during training and enabling self-debugging through execution feedback.

\noindent\textbf{Multi-Task Learning.}
Multi-task learning improves generalization through shared representations~\cite{mtl-overview}. The \textit{Cocktail} effect~\cite{data-mixing} produces synergistic gains from complementary tasks. FLAN~\cite{wei2022finetuned} and T0~\cite{sanh2022multitask} mix instruction formats, while CodeT5+~\cite{wang2023codet5plus} jointly trains on code understanding and generation. Our \textit{Cocktail} dataset uniquely combines three supervision signals: (i) formula-to-Prolog translation for symbolic grounding, (ii) natural language comments for semantic alignment, and (iii) program execution for numeric verification, mixing reasoning modalities rather than task formats.

\noindent\textbf{Program Synthesis and Error Correction.}
Neural program synthesis methods include Codex~\cite{chen2021codex}, AlphaCode~\cite{li2022alphacode}, and CodeLlama~\cite{codellama}. Self-Repair~\cite{chen2023teaching} trains on error-correction pairs, while Reflexion~\cite{shinn2023reflexion} uses verbal feedback for iterative improvement. These methods target general-purpose programming rather than formal reasoning. We extend execution-guided decoding by leveraging Prolog's error taxonomy to guide systematic self-correction.

\noindent\textbf{Gaps in Prior Work.}
Table~\ref{tab:related_comparison} situates NeuroProlog along five dimensions that collectively define the design space of neurosymbolic mathematical reasoning systems.

\begin{table}[t]
\centering
\caption{Comparison of neurosymbolic reasoning systems across five design dimensions. \checkmark~=~fully supported; P~=~partial; $\times$~=~absent.}
\label{tab:related_comparison}
\renewcommand{\arraystretch}{1.15}
\setlength{\tabcolsep}{3pt}
\scriptsize
\begin{tabular}{@{}lccccccc|c@{}}
\toprule
\textbf{Dimension}
  & \rotatebox{65}{\textbf{PAL/PoT}}
  & \rotatebox{65}{\textbf{GSM8K-P}}
  & \rotatebox{65}{\textbf{LogicLM}}
  & \rotatebox{65}{\textbf{LINC}}
  & \rotatebox{65}{\textbf{Self-Dbg}}
  & \rotatebox{65}{\textbf{Reflexion}}
  & \rotatebox{65}{\textbf{ProSLM}}
  & \rotatebox{65}{\textbf{Ours}} \\
\midrule
KB grounding (train)     & $\times$ & $\times$ & $\times$ & $\times$ & $\times$ & $\times$ & P          & \checkmark \\
Cross-task transfer      & $\times$ & $\times$ & $\times$ & $\times$ & $\times$ & $\times$ & $\times$   & \checkmark \\
5-class error taxonomy   & $\times$ & $\times$ & $\times$ & $\times$ & $\times$ & $\times$ & $\times$   & \checkmark \\
Execution feedback       & $\times$ & $\times$ & P        & $\times$ & \checkmark & \checkmark & $\times$ & \checkmark \\
Scale-dependent analysis & $\times$ & $\times$ & $\times$ & $\times$ & $\times$ & $\times$ & $\times$   & \checkmark \\
\bottomrule
\end{tabular}
\end{table}

No prior work jointly trains on formula-to-KB translation or quantifies cross-task transfer between declarative and procedural symbolic objectives---gaps NeuroProlog addresses, with the \textit{Cocktail} Effect validated empirically in Section~\ref{subsec:ablation}.
\section{Methodology}
\label{sec:methodology}

We present \textsc{NeuroProlog}, a neurosymbolic framework that enforces executable mathematical reasoning through multi-task fine-tuning with Prolog-based symbolic supervision. Our approach consists of three core components: (1) a unified training corpus combining declarative mathematical knowledge with procedural problem-solving demonstrations (Section~\ref{subsec:dataset}), (2) a multi-task \textit{Cocktail} training objective that induces cross-task transfer through shared symbolic representations (Section~\ref{subsec:training}), and (3) an execution-guided decoding pipeline with iterative error feedback (Section~\ref{subsec:inference}).

\subsection{Dataset Construction}
\label{subsec:dataset}

We construct a unified training corpus designed to enable neurosymbolic mathematical reasoning through Prolog-based representations. Our dataset comprises two complementary components: a \textbf{Mathematical Knowledge Base (KB)} providing declarative symbolic grounding, and a \textbf{Problem-Solving Dataset} demonstrating procedural application of mathematical concepts.

\subsubsection*{Mathematical Knowledge Base}
The KB comprises 200 carefully constructed entries formalizing fundamental mathematical concepts as executable Prolog predicates. Each entry follows a standardized instruction-input-output format:

\begin{myexample}{Knowledge Base Entry}
\textbf{Instruction:} [KNOWLEDGE] Generate Prolog code that encodes the given mathematical concept

\textbf{Input:} Natural language description of a mathematical formula, concept, or inference rule with semantic explanation and worked examples

\textbf{Output:} Executable Prolog program implementing the concept with module imports, natural language comments, type-safe predicate definitions, helper predicates for compositional reasoning, and a standardized \texttt{solve/1} interface.
\end{myexample}

\noindent\textbf{Domain Coverage.} The KB entries span 11+ mathematical domains: Basic Statistics (22 entries, 11.0\%), Ratio \& Proportion (17, 8.5\%), Percentage Calculations (17, 8.5\%), Number Theory (11, 5.5\%), Propositional Logic (9, 4.5\%), Boolean Operations (7, 3.5\%), Geometry (12, 6.0\%), Linear Equations (6, 3.0\%), Sequences \& Series (3, 1.5\%), and mixed topics (96, 48.0\%).

\noindent\textbf{Semantic Alignment Through Comments.} A critical design choice distinguishes our KB from conventional code datasets: every Prolog predicate includes natural language comments that explain its mathematical semantics, enabling the model to bridge informal and formal reasoning.

\subsubsection*{Problem-Solving Dataset}
The second component consists of 310 problem-solving examples based on the KB, together with 7476 entries from GSM8K-Prolog~\cite{yang2024arithmetic} with executable Prolog programs:

\begin{myexample}{Problem-Solving Entry Format}
\textbf{Instruction:} [SOLVE] Generate correct Prolog code that solves the given math problem

\textbf{Input:} Mathematical word problem with numeric constraints and a question

\textbf{Output:} Complete Prolog program that encodes problem constraints as predicates, implements the reasoning chain, and produces the numeric answer via \texttt{solve(Result)}.
\end{myexample}

\noindent\textbf{Quality Assurance.} Each entry underwent multi-stage validation: (1) syntactic verification via SWI-Prolog parsing; (2) execution testing of each \texttt{solve/1} predicate; (3) semantic review by domain experts and researchers.

\subsection{Multi-Task \textit{Cocktail} Training}
\label{subsec:training}

We formulate neurosymbolic mathematical reasoning as a multi-task learning problem where models jointly acquire complementary capabilities through shared symbolic representations.

\noindent\textbf{Task Formulation.}
Let $\mathcal{D}_{\text{KB}} = \{(x_i^{\text{kb}}, y_i^{\text{kb}})\}_{i=1}^{200}$ denote the Knowledge Base dataset, where $x_i^{\text{kb}}$ is a natural language description and $y_i^{\text{kb}}$ is its Prolog implementation. Let $\mathcal{D}_{\text{SOLVE}} = \{(x_j^{\text{solve}}, y_j^{\text{solve}})\}_{j=1}^{7786}$ denote the Problem-Solving dataset, where $x_j^{\text{solve}}$ is a word problem and $y_j^{\text{solve}}$ is its executable Prolog program. We train $f_\theta$ to minimize:
\begin{equation}
\mathcal{L}_{\text{cocktail}}(\theta)
=
\lambda_{\text{kb}} \mathcal{L}_{\text{KB}}(\theta)
+
\lambda_{\text{solve}} \mathcal{L}_{\text{SOLVE}}(\theta),
\label{eq:cocktail}
\end{equation}
with each task-specific loss being the standard causal language modeling objective:
\begin{align}
\mathcal{L}_{\text{KB}}(\theta)
&=
-\mathbb{E}_{(x,y)\sim \mathcal{D}_{\text{KB}}}
\left[
\sum_{t=1}^{|y|}
\log p_\theta(y_t \mid x, y_{<t})
\right], \label{eq:loss-kb}\\
\mathcal{L}_{\text{SOLVE}}(\theta)
&=
-\mathbb{E}_{(x,y)\sim \mathcal{D}_{\text{SOLVE}}}
\left[
\sum_{t=1}^{|y|}
\log p_\theta(y_t \mid x, y_{<t})
\right]. \label{eq:loss-solve}
\end{align}

Task weights satisfy $\lambda_{\text{kb}} + \lambda_{\text{solve}} = 1$. Verified answers $a_j$ are used only during evaluation to identify logically incorrect executable programs; they are not used during training.

\noindent\textbf{Training Protocol.}
We fine-tune using LoRA (without quantization), injecting low-rank adapters into attention layers while freezing pretrained weights. Both KB and SOLVE entries are formatted with task-type prefixes (\texttt{[KNOWLEDGE]} vs.~\texttt{[SOLVE]}) to enable task-conditional generation. All models share a common optimization setup (3 epochs, cosine learning rate schedule with 8\% warmup, gradient clipping at 1.0, bfloat16 precision). LoRA rank, learning rate, and regularization are tuned per model scale, with gradient accumulation yielding effective batch sizes of 32--64.

\noindent\textbf{\textit{Cocktail} Effect Hypothesis.}
The \textit{Cocktail} effect~\cite{brief2024mixing} suggests complementary multi-task training yields synergistic gains. We hypothesize three transfer mechanisms: (1) \textit{Compositional Reuse} (KB $\rightarrow$ SOLVE): KB predicates become reusable building blocks for SOLVE programs; (2) \textit{Contextual Grounding} (SOLVE $\rightarrow$ KB): problem-solving provides usage patterns reinforcing abstract formula semantics; (3) \textit{Unified Type System}: joint training on Prolog syntax and CLP(Q) constraints induces shared type-safe representations.

\subsection{Execution-Guided Decoding with Error Feedback}
\label{subsec:inference}

At inference time, we introduce an iterative pipeline that leverages Prolog's error diagnostics to enable self-debugging through execution-aware generation.

\subsubsection{Pipeline Architecture}

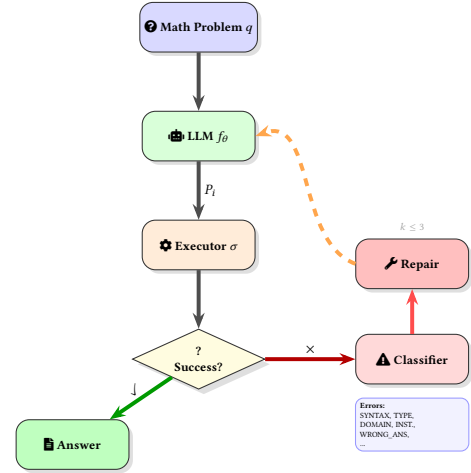
\begin{figure}[t]
\centering
\begin{tikzpicture}[
    scale=0.6,
    transform shape,
    box/.style={
        rectangle, draw=black!80, rounded corners=5pt,
        minimum width=2.5cm, minimum height=1.1cm,
        align=center, font=\small\bfseries,
        drop shadow={opacity=0.25, shadow xshift=1.5pt, shadow yshift=-1.5pt}
    },
    decision/.style={
        diamond, draw=black!80, aspect=2.2,
        minimum width=2cm, align=center, font=\small\bfseries,
        drop shadow={opacity=0.25}
    },
    arrow/.style={->, line width=1.5pt, -{Stealth[length=6pt]}}
]
\node[box, fill=blue!15] (q) {Q: Math Problem $q$};
\node[box, fill=green!15, below=1.3cm of q] (llm) {LLM: LLM $f_\theta$};
\node[box, fill=orange!15, below=1.3cm of llm] (exec) {Exec: Executor $\sigma$};
\node[decision, fill=yellow!15, below=1.3cm of exec] (check) {?\\Success?};
\node[box, fill=green!25, below left=1cm and 0.8cm of check] (out){Out: Answer};
\node[box, fill=red!15, right=2cm of check] (err) {Err: Classifier};
\node[box, fill=red!25,above=1cm of err] (fix) {Fix: Repair};

\draw[arrow, draw=black!70] (q) -- (llm);
\draw[arrow, draw=black!70] (llm) -- node[right] {$P_i$} (exec);
\draw[arrow, draw=black!70] (exec) -- (check);
\draw[arrow, draw=green!60!black] (check) -- node[above, sloped] {\checkmark} (out);
\draw[arrow, draw=red!70!black] (check) -- node[above] {$\times$} (err);
\draw[arrow, draw=red!70] (err) -- (fix);
\draw[arrow, draw=orange!70, dashed] (fix.west) to[out=160, in=20] (llm.east);

\node[font=\scriptsize, text=gray!70, above=2pt of fix] {$k \leq 3$};

\node[draw=blue!40, fill=blue!5, rounded corners=3pt,
      font=\tiny, text width=2.3cm, align=left,
      below=0.3cm of err] {
    \textbf{Errors:}\\
    SYNTAX, TYPE,\\
    DOMAIN, INST.,\\
    WRONG\_ANS, ...
};
\end{tikzpicture}
\caption{NeuroProlog execution-guided decoding pipeline with error feedback.}
\label{fig:architecture}
\end{figure}

Fig.~\ref{fig:architecture} illustrates our four-phase pipeline. Given math problem $q$, fine-tuned LLM $f_\theta$ generates initial Prolog program $P_0$. The SWI-Prolog interpreter $\sigma$ executes $P_0$, producing three possible outcomes: (1) Success with correct answer (terminate); (2) Success with wrong answer (classify as \texttt{LOGICAL\_ERROR}); (3) Error with diagnostic $e$ (classify via taxonomy). On failure, we generate targeted repair prompt $\mathcal{P}_e$ containing the original problem, failed program, error type, and fix instructions. The model produces corrected program $P_{i+1} = f_\theta(\mathcal{P}_e)$, repeating up to $k=3$ iterations.

\subsubsection{Error Taxonomy and Repair Strategies}
\label{subsubsec:error-taxonomy}

We distinguish five error classes from SWI-Prolog diagnostics: \textit{Syntax} (malformed tokens/brackets), \textit{Type} (arithmetic on non-numeric terms), \textit{Domain} (runtime violations such as division by zero or negative factorial), \textit{Instantiation} (variables used before binding), and \textit{Logical} (\texttt{WRONG\_ANSWER}: program executes but $\sigma(P) = a' \neq a^*$). For each error class, a targeted repair prompt is applied to address the corresponding diagnostic category, with class-specific diagnostic codes and prompt templates used to guide the repair process.
\subsection{Evaluation Metrics}
\label{subsec:metrics}

We measure performance through five complementary metrics: \textbf{Accuracy} (final correct answers after $k \leq 3$ iterations); \textbf{Executability} (fraction with at least one valid Prolog execution); \textbf{initial Success} (solved on attempt 1); \textbf{Correction Rate} (fraction of initial failures eventually repaired); and \textbf{Average Iterations} (mean execution attempts per problem).

\section{Experiments}
\label{sec:experiments}

We conduct comprehensive evaluation of NeuroProlog across four LLMs spanning a 10$\times$ parameter range (3B--32B).

\subsection{Experimental Setup}
\label{sec:setup}

\noindent\textbf{Models.} We evaluate four models: (i)~\textbf{Qwen2.5-Coder-32B-Instruct} (code-specialized), (ii)~\textbf{GPT-OSS-20B} (general-purpose), (iii)~\textbf{Qwen3-8B} (mid-scale general), and (iv)~\textbf{Llama-3.2-3B-Instruct} (compact instruction-tuned).

\noindent\textbf{Training Configurations.} For each model we compare three settings: \textbf{Base} (zero-shot), \textbf{Prolog~FT} (fine-tuned on GSM8K-Prolog), and \textbf{\textit{Cocktail}~FT} (fine-tuned on our unified KB+SOLVE dataset per Eq.~\ref{eq:cocktail}). Both fine-tuned configurations use identical LoRA hyperparameters and train for the same number of epochs.

\noindent\textbf{Decoding and Evaluation.} We use greedy decoding (temperature$=0$) for deterministic results. The pipeline runs with $k{=}3$ maximum correction iterations on the full GSM8K test set ($n{=}1{,}319$). Significance is assessed via McNemar's test.

\subsection{Overall Performance}
\label{sec:overall}

\begin{figure}[t]
\centering
\includegraphics[width=\columnwidth]{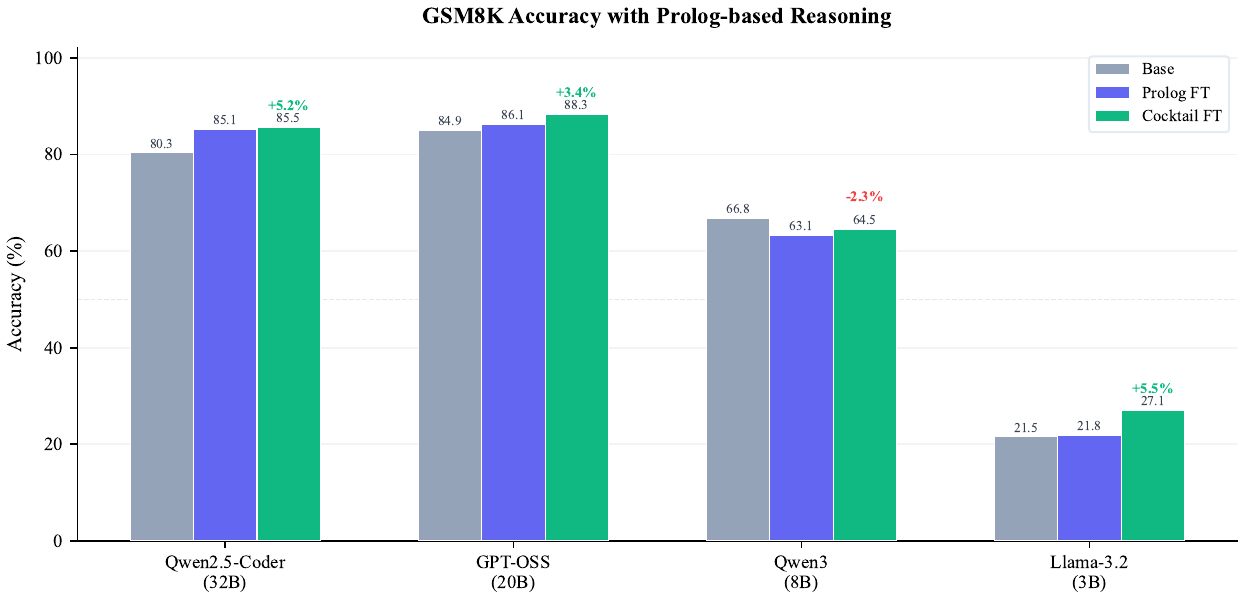}
\caption{GSM8K accuracy across four models and three configurations. Annotations show \textit{Cocktail} FT improvement over the base model. Qwen3-8B is the only model where fine-tuning decreases accuracy, revealing a generation--correction trade-off.}
\label{fig:accuracy}
\end{figure}

Table~\ref{tab:main_results} presents evaluation across all twelve model-configuration pairs. Fig.~\ref{fig:accuracy} summarizes accuracy visually.

\begin{table*}[t]
\centering
\caption{GSM8K evaluation with Prolog-based code generation. \textbf{Bold} = best per model; \underline{underline} = overall best. $\Delta$ rows show absolute change from the base model. $\dagger$: $p < 0.05$; $\ddagger$: $p < 0.01$ by McNemar's test against the same-model base configuration.}
\label{tab:main_results}
\small
\begin{tabular}{@{}llcccccc@{}}
\toprule
\textbf{Model} & \textbf{Config} & \textbf{Acc.~(\%)} & \textbf{Exec.~(\%)} & \textbf{Initial Exec~(\%)} & \textbf{Corr.~(\%)} & \textbf{Avg Iter} & \textbf{Sig.} \\
\midrule
\multicolumn{8}{l}{\cellcolor{gray!5}\textit{Large-Scale Code Model (32B)}} \\
Qwen2.5-Coder-32B & Base        & 80.29 & 97.04 & 96.4 & 17.0 & 1.07 & --- \\
               & Prolog FT   & 85.14 & 99.24 & \textbf{98.7} & 41.2 & \textbf{1.02} & $\ddagger$ \\
               & Cocktail FT & \textbf{85.52} & \textbf{99.32} & 90.7 & \textbf{92.7} & 1.11 & $\ddagger$ \\
               & $\Delta_{\text{Prolog}}$   & +4.85 & +2.20 & +2.3  & +24.2 & $-$0.05 & \\
               & $\Delta_{\text{Cocktail}}$ & \textbf{+5.23} & +2.28 & $-$5.7 & +75.7 & +0.04 & \\
\midrule
\multicolumn{8}{l}{\cellcolor{gray!5}\textit{General-Purpose Model (20B)}} \\
GPT-OSS-20B & Base        & 84.91 & 89.99 & 45.0 & 81.8 & 1.75 & --- \\
             & Prolog FT   & 86.12 & 91.00 & 55.0 & 69.2 & 1.61 & $\dagger$ \\
             & Cocktail FT & \underline{\textbf{88.34}} & \textbf{93.00} & \textbf{60.0} & 70.8 & \textbf{1.52} & $\ddagger$ \\
             & $\Delta_{\text{Prolog}}$   & +1.21 & +1.01 & +10.0 & $-$12.6 & $-$0.14 & \\
             & $\Delta_{\text{Cocktail}}$ & \textbf{+3.43} & +3.01 & +15.0 & $-$11.0 & $-$0.23 & \\
\midrule
\multicolumn{8}{l}{\cellcolor{gray!5}\textit{Mid-Scale General Model (8B)}} \\
Qwen3-8B & Base        & \textbf{66.79} & 72.93 & 7.7 & \textbf{70.7} & 2.43 & --- \\
          & Prolog FT   & 63.15 & 75.96 & 69.2 & 27.7 & 1.60 & $\dagger$ \\
          & Cocktail FT & 64.51 & \textbf{77.10} & \textbf{69.8} & 24.1 & \textbf{1.54} & \\
          & $\Delta_{\text{Prolog}}$   & $-$3.64 & +3.03 & +61.5 & $-$43.0 & $-$0.83 & \\
          & $\Delta_{\text{Cocktail}}$ & $-$2.28 & +4.17 & +62.1 & $-$46.6 & $-$0.89 & \\
\midrule
\multicolumn{8}{l}{\cellcolor{gray!5}\textit{Small-Scale Model (3B)}} \\
Llama-3B-Instruct& Base        & 21.53 & \textbf{66.49} & 47.5 & 36.2 & 1.90 & --- \\
              & Prolog FT   & 21.83 & 42.08 & 23.4 & 24.4 & 2.36 & \\
              & Cocktail FT & \textbf{27.07} & 57.92 & \textbf{53.8} & 8.9 & \textbf{1.89} & $\dagger$ \\
              & $\Delta_{\text{Prolog}}$   & +0.30 & $-$24.41 & $-$24.1 & $-$11.8 & +0.46 & \\
              & $\Delta_{\text{Cocktail}}$ & \textbf{+5.54} & $-$8.57 & +6.3 & $-$27.3 & $-$0.01 & \\
\bottomrule
\end{tabular}
\end{table*}

\noindent\textbf{\textit{Cocktail} training yields significant gains for three of four models.}
As shown in Fig.~\ref{fig:accuracy}, \textit{Cocktail} training improves accuracy by +5.23\% ($p < 0.01$, Qwen-32B), +3.43\% ($p < 0.01$, GPT-OSS-20B), and +5.54\% ($p < 0.05$, Llama-3B). In each case, \textit{Cocktail} FT also outperforms single-task Prolog FT (e.g., by +2.22\% for GPT-OSS). The strongest overall result---88.34\% by GPT-OSS-20B \textit{Cocktail} FT---demonstrates that multi-task neurosymbolic training can unlock performance in general-purpose models surpassing larger code-specialized architectures.

\noindent\textbf{Qwen3-8B reveals a generation--correction trade-off.}
Qwen3-8B is the only model where fine-tuning \emph{decreases} accuracy. The base model achieves 66.79\% by compensating for poor initial generation (7.7\% initial execution) through strong iterative repair (70.7\% correction). Fine-tuning inverts this profile: initial execution success leaps to 69.8\% but correction collapses to 24.1\%.

\noindent\textbf{Generalization to MATH-500}
\label{sec:math500}

To assess out-of-distribution generalization, we evaluate on a 239-problem subset of MATH-500 covering problems with real numeric answers; problems requiring imaginary numbers, symbolic equation generation, or geometric constructions are excluded due to SWI-Prolog CLP(Q) scope. The evaluation protocol is held constant with GSM8K: identical base models, LoRA hyperparameters, and three-condition structure (Base, Prolog FT, \textit{Cocktail} FT). No MATH-500 data was observed during training.

\begin{table}[t]
\centering
\caption{\textit{Cocktail} FT accuracy (\%) on MATH-500 subset (239 problems). $\Delta$ denotes change from Base.}
\label{tab:math500}
\small
\begin{tabular}{@{}lcccc@{}}
\toprule
\textbf{Model} & \textbf{Base} & \textbf{\textit{Cocktail} FT} & \textbf{$\Delta$} \\
\midrule
Qwen2.5-Coder-32B  & 40.17 & 42.68 & \textbf{+2.51} \\
GPT-OSS-20B        & 49.37 & 51.46 & \textbf{+2.09} \\
Llama-3B-Instruct  &  9.10 & 10.04 & \textbf{+0.94} \\
Qwen3-8B           & 30.96 & 21.34 & $-9.62$ \\
\bottomrule
\end{tabular}
\end{table}

The combined KB+SOLVE training configuration improves accuracy over the corresponding single-task Prolog FT configuration on three of four models, indicating that incorporating complementary symbolic supervision can provide additional benefits.

\subsection{Effect of Model Scale}
\label{sec:scale}

\begin{figure}[t]
\centering
\includegraphics[width=0.8\columnwidth]{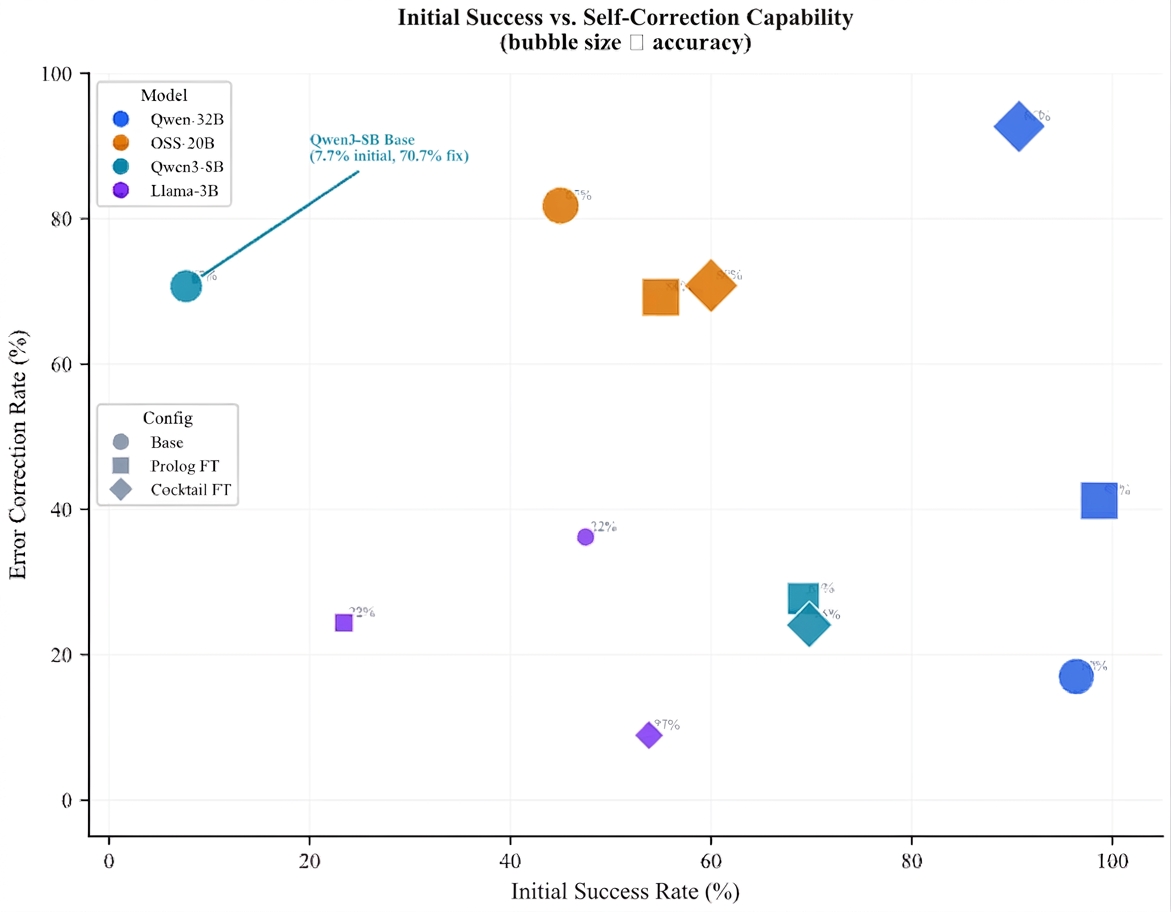}
\caption{Initial success vs.\ correction rate across all 12 configurations, revealing four distinct behavioral regimes.}
\label{fig:scatter}
\end{figure}

Fig.~\ref{fig:scatter} plots initial execution success against correction rate for all twelve configurations, revealing four distinct regimes.

\noindent\textbf{Large code models (32B): Clean generation, emergent correction.}
Qwen-32B achieves near-perfect initial success (96.4\%) but weak correction (17.0\%). \textit{Cocktail} training induces a qualitative shift: initial success decreases to 90.7\% while correction improves to 92.7\% ($p<0.01$).

\noindent\textbf{Medium general models (20B): Error tolerance.}
GPT-OSS-20B compensates for moderate initial success (45.0\%) with strong correction (81.8\%). \textit{Cocktail} training shifts toward first-attempt resolution ($45\%\rightarrow60\%$) while maintaining correction quality.

\noindent\textbf{Mid-scale models (8B): Generation--correction trade-off.}
Qwen3-8B Base has low initial success (7.7\%) but strong correction (70.7\%). Fine-tuning boosts initial success by $9\times$ but sharply reduces correction, lowering overall accuracy ($p=0.08$).

\noindent\textbf{Small models (3B): Capacity-limited.}
Llama-3B-Instruct remains in the lower-left region regardless of configuration. The 27.07\% \textit{Cocktail} FT accuracy ceiling confirms 3B parameters are insufficient for reliable compositional Prolog generation.

\subsection{Execution-Guided Error Recovery}
\label{sec:recovery}

\begin{figure}[t]
\centering
\includegraphics[width=\columnwidth]{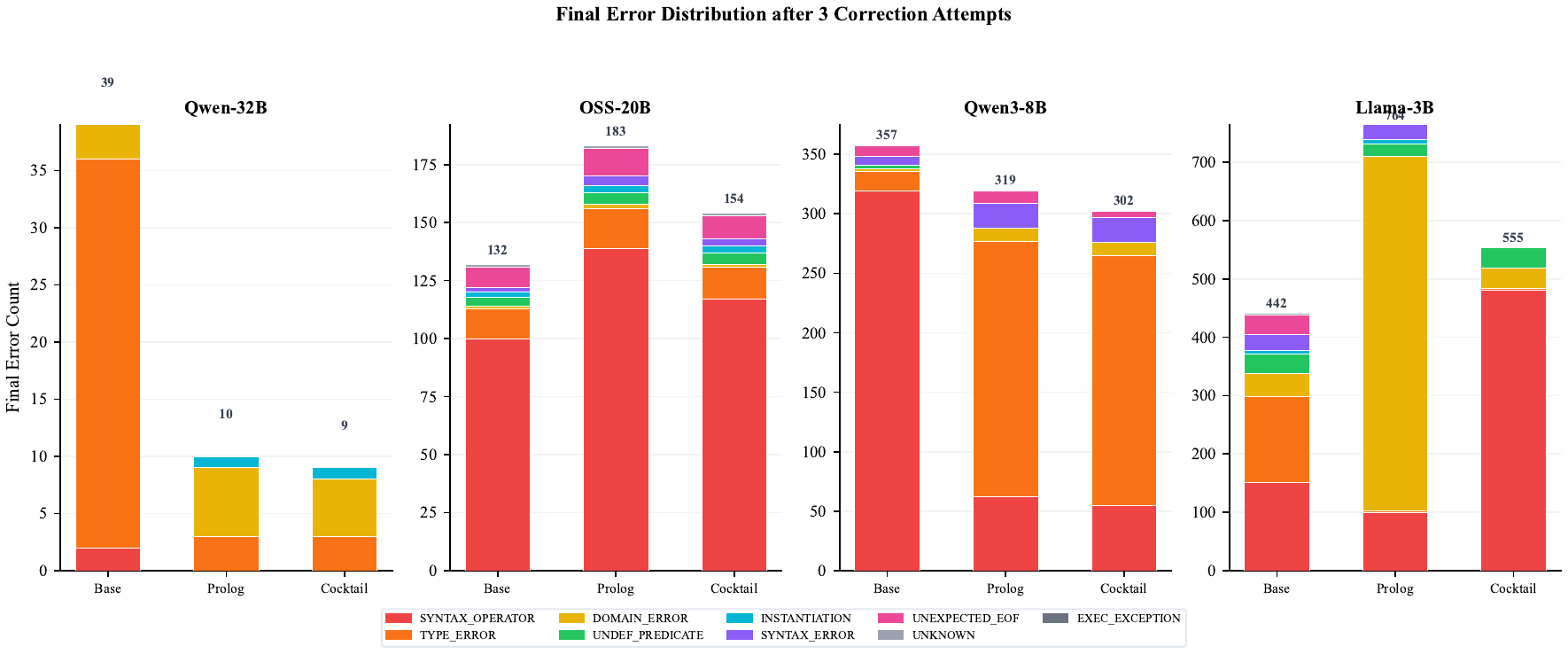}
\caption{Final error distribution after $k{=}3$ correction attempts, by Prolog error type.}
\label{fig:errors}
\end{figure}

\noindent\textbf{\textit{Cocktail} FT trades inital success perfection for robust correction (Qwen-32B).}
\textit{Cocktail} training reduces Qwen-32B's initial success accuracy from 96.4\% to 90.7\% but enables 92.7\% error repair, yielding only 9 final failures versus 39 for the base model (77\% reduction).

\noindent\textbf{GPT-OSS-20B: Monotonic pipeline improvement.}
Fine-tuning monotonically improves all pipeline metrics: initial success ($45\% \to 60\%$), average iterations ($1.75 \to 1.52$), and accuracy ($84.91\% \to 88.34\%$). The \textit{Cocktail} configuration requires 13.1\% fewer total iterations than the base while achieving +3.43\% higher accuracy.

\noindent\textbf{Correction degrades at sub-10B scale.}
Qwen3-8B and Llama-3B-Instruct exhibit sharp declines in correction rates after fine-tuning (70.7\%$\rightarrow$24.1\% and 36.2\%$\rightarrow$8.9\%, respectively), indicating that models below $\sim$10B parameters lack sufficient capacity to jointly learn code generation and execution-based error diagnosis.

\subsection{Error Analysis}
\label{sec:error_analysis}

\noindent\textbf{Scale-dependent error shift---the key mechanistic finding.}
The most informative result from Fig.~\ref{fig:errors} is the \emph{opposite} error transformations at different model capacities. At 32B, base model errors are 87.2\% \texttt{TYPE\_ERROR}---semantic failures fixable only 12\% of the time. After \textit{Cocktail} training, errors shift to 55.6\% \texttt{DOMAIN\_ERROR} (repaired at 96\%). Total errors drop from 39 to 9.

At 8B, the \emph{inverse} occurs: Qwen3-8B base errors are 89.4\% \texttt{SYNTAX\_OPERATOR\_EXPECTED}. After \textit{Cocktail} training, the dominant error becomes 69.5\% \texttt{TYPE\_ERROR} with only 5\% fixability. Fine-tuning successfully teaches Prolog syntax but \emph{introduces} the same semantic failure that the 32B model resolves, suggesting semantic type understanding requires capacity beyond 8B parameters.

\subsection{Pipeline Efficiency}
\label{sec:efficiency}

\begin{figure}[ht]
\centering
\includegraphics[width=0.75\columnwidth]{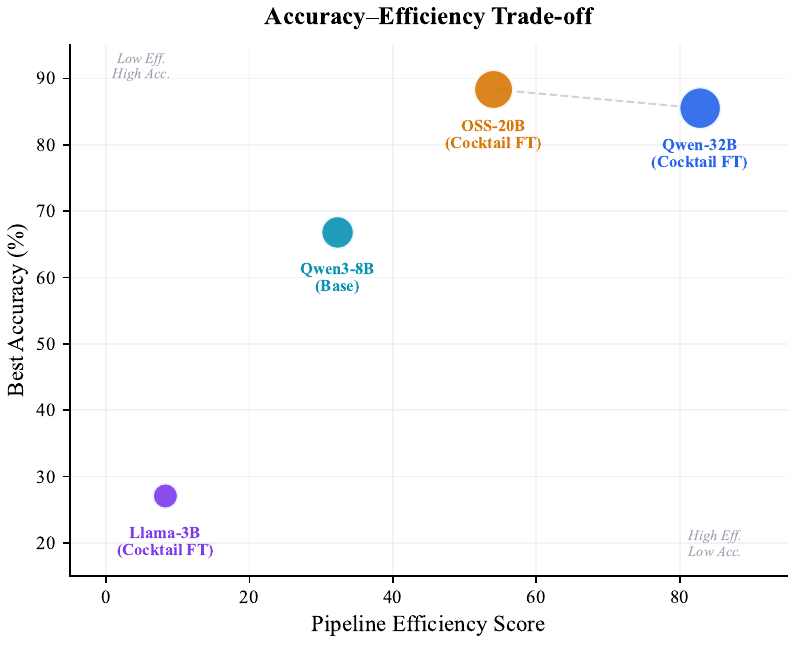}
\caption{Accuracy vs.\ Efficiency tradeoff across all configurations.}
\label{fig:efficiency}
\end{figure}

We define pipeline efficiency as:
\begin{equation}
\text{Efficiency} = \frac{\text{Accuracy} \times \text{Execution Rate}}{\text{Average Iterations}}.
\label{eq:efficiency}
\end{equation}
Qwen3-8B shows the largest efficiency gain (+61.5\%) despite accuracy loss, as reduced iterations ($2.43 \rightarrow 1.54$) and improved executability outweigh accuracy decline for latency-sensitive applications. Fig.~\ref{fig:efficiency} identifies two Pareto-optimal points: GPT-OSS-20B \textit{Cocktail} for accuracy (88.34\%) and Qwen-32B Prolog FT for efficiency.

\subsection{Ablation Studies}
\label{subsec:ablation}

\begin{table}[h]
\centering
\small
\caption{Effect of Additional Symbolic Supervision}
\label{tab:kb-ablation}
\begin{tabular}{@{}lrrrr@{}}
\toprule
\textbf{Model} & \textbf{Prolog FT} & \textbf{\textit{Cocktail} FT} & \textbf{$\Delta$Acc.} & \textbf{$\Delta$Exec.} \\
\midrule
Qwen-32B       & 85.14 & 85.52 & +0.38 & +0.08 \\
GPT-OSS-20B    & 86.12 & 88.34 & +2.22 & +2.00 \\
Qwen3-8B       & 63.15 & 64.51 & +1.36 & +1.14 \\
Llama-3B       & 21.83 & 27.07 & +5.24 & +15.84 \\
\midrule
\textbf{Mean}  & 64.06 & 66.36 & \textbf{+2.30} & \textbf{+4.77} \\
\bottomrule
\end{tabular}
\end{table}

\noindent\textbf{KB Integration.} Table~\ref{tab:kb-ablation} compares the single-task Prolog FT configuration with the combined KB+SOLVE configuration. The addition of KB supervision improves accuracy across all four models, with gains ranging from +0.38\% to +5.24\%.

\noindent\textbf{Execution-Guided Decoding.} Taxonomy-based feedback adds +4.82\% over greedy decoding. Qwen-32B \textit{Cocktail} achieves 90.7\% initial success success, with 92.7\% correction rate demonstrating effective execution-guided repair without correction-specific training.

\begin{table}[h]
\centering
\small
\caption{Effect of execution-guided repair on Qwen-32B performance.}
\label{tab:synergy}
\begin{tabular}{@{}lccc@{}}
\toprule
 & \textbf{Initial success(\%)} & \textbf{Final accuracy(\%)} & \textbf{$\Delta$} \\
\midrule
Prolog FT   & 98.7 & 85.14 & $-13.6$\\
\textit{Cocktail} FT & 90.7 & 85.52 & $-5.2$ \\
\midrule
\textbf{$\Delta$ (KB)} & $-8.0$ & \textbf{+0.38} & --- \\
\bottomrule
\end{tabular}
\end{table}

Table~\ref{tab:synergy} compares initial generation performance with final accuracy after execution-guided repair for Qwen-32B. The results show that the \textit{Cocktail} configuration exhibits a smaller gap between initial and final performance than Prolog FT. This comparison characterizes the interaction between generation quality and subsequent repair, but does not establish a causal effect of KB supervision on repair.

\section{Conclusion}
\label{sec:conclusion}

We presented NeuroProlog, a neurosymbolic framework that combines multi-task symbolic supervision with execution-guided program repair for mathematical reasoning. The combined KB+SOLVE training improves GSM8K accuracy over single-task Prolog fine-tuning for the evaluated models, with gains of up to 5.54\%, while MATH-500 results indicate generalization to unseen problems. Error analysis reveals scale-associated differences in repair behavior, with larger models showing greater improvements in handling semantic errors. These results demonstrate the potential of combining symbolic knowledge and execution feedback for improving LLM-based mathematical reasoning.

Limitations and future work include scaling the KB to 1000+ entries for advanced mathematics, comparative studies across SMT solvers and Lean, controlled scaling experiments (1B--70B), and evaluation on MATH and TheoremQA benchmarks.

\bibliographystyle{unsrt}
\bibliography{references}

@article{gsm8k,
  author  = {Cobbe, K. and Kosaraju, V. and Bavarian, M. and Chen, M. and Jun, H. and Kaiser, L. and Plappert, M. and Tworek, J. and Hilton, J. and Nakano, R. and Hesse, C. and Schulman, J.},
  title   = {Training Verifiers to Solve Math Word Problems},
  journal = {arXiv preprint arXiv:2110.14168},
  year    = {2021}
}

@inproceedings{wei2022cot,
  author    = {Wei, J. and Wang, X. and Schuurmans, D. and Bosma, M. and Ichter, B. and Xia, F. and Chi, E. H. and Le, Q. V. and Zhou, D.},
  title     = {Chain-of-Thought Prompting Elicits Reasoning in Large Language Models},
  booktitle = {Advances in Neural Information Processing Systems},
  volume    = {35},
  pages     = {24824--24837},
  year      = {2022}
}

@article{poT2023,
  author  = {Chen, W. and Ma, X. and Wang, X. and Cohen, W. W.},
  title   = {Program of Thoughts Prompting: Disentangling Computation from Reasoning for Numerical Reasoning Tasks},
  journal = {Transactions on Machine Learning Research},
  year    = {2023}
}

@inproceedings{lyu2023faithful,
  author    = {Lyu, Q. and Havaldar, S. and Stein, A. and Zhang, L. and Rao, D. and Wong, E. and Apidianaki, M. and Callison-Burch, C.},
  title     = {Faithful Chain-of-Thought Reasoning},
  booktitle = {Proceedings of the 13th International Joint Conference on Natural Language Processing and the 3rd Conference of the Asia-Pacific Chapter of the Association for Computational Linguistics (Volume 1: Long Papers)},
  pages     = {305--329},
  address   = {Nusa Dua, Bali},
  publisher = {Association for Computational Linguistics},
  year      = {2023},
  doi       = {10.18653/v1/2023.ijcnlp-main.20}
}

@article{valmeekam2022plan,
  author  = {Valmeekam, K. and Olmo, A. and Sreedharan, S. and Kambhampati, S.},
  title   = {Large Language Models Still Can't Plan (A Benchmark for LLMs on Planning and Reasoning about Change)},
  journal = {arXiv preprint arXiv:2206.10498},
  year    = {2022}
}

@inproceedings{creswell2022selection,
  author    = {Creswell, A. and Shanahan, M. and Higgins, I.},
  title     = {Selection-Inference: Exploiting Large Language Models for Interpretable Logical Reasoning},
  booktitle = {International Conference on Learning Representations},
  year      = {2023}
}

@article{singh2026verge,
  author  = {Singh, V. and Cassel, D. and Weir, N. and Feng, N. and Bayless, S.},
  title   = {{VERGE}: Formal Refinement and Guidance Engine for Verifiable {LLM} Reasoning},
  journal = {arXiv preprint arXiv:2601.20055},
  year    = {2026}
}

@inproceedings{chen2025neurosymbolic,
  author    = {Chen, M. K.},
  title     = {A Comparative Study of Neurosymbolic AI Approaches to Interpretable Logical Reasoning},
  booktitle = {Proceedings of NeSy 2025},
  year      = {2025}
}

@inproceedings{logic-lm,
  author    = {Pan, L. and Albalak, A. and Wang, X. and Wang, W.},
  title     = {{Logic-LM}: Empowering Large Language Models with Symbolic Solvers for Faithful Logical Reasoning},
  booktitle = {Findings of the Association for Computational Linguistics: EMNLP 2023},
  pages     = {3806--3824},
  address   = {Singapore},
  publisher = {Association for Computational Linguistics},
  year      = {2023},
  doi       = {10.18653/v1/2023.findings-emnlp.248}
}

@inproceedings{linc,
  author    = {Olausson, T. X. and Gu, A. and Lipkin, B. and Zhang, C. E. and Solar-Lezama, A. and Tenenbaum, J. B. and Levy, R.},
  title     = {{LINC}: A Neurosymbolic Approach for Logical Reasoning by Combining Language Models with First-Order Logic Provers},
  booktitle = {Proceedings of the 2023 Conference on Empirical Methods in Natural Language Processing},
  pages     = {5153--5176},
  publisher = {Association for Computational Linguistics},
  year      = {2023},
  doi       = {10.18653/v1/2023.emnlp-main.313}
}

@inproceedings{kojima2022large,
  author    = {Kojima, T. and Gu, S. S. and Reid, M. and Matsuo, Y. and Iwasawa, Y.},
  title     = {Large Language Models Are Zero-Shot Reasoners},
  booktitle = {Advances in Neural Information Processing Systems},
  volume    = {35},
  pages     = {22199--22213},
  year      = {2022}
}

@inproceedings{wang2023selfconsistency,
  author    = {Wang, X. and Wei, J. and Schuurmans, D. and Le, Q. V. and Chi, E. H. and Narang, S. and Chowdhery, A. and Zhou, D.},
  title     = {Self-Consistency Improves Chain of Thought Reasoning in Language Models},
  booktitle = {International Conference on Learning Representations},
  year      = {2023}
}

@inproceedings{gao2023pal,
  author    = {Gao, L. and Madaan, A. and Zhou, S. and Alon, U. and Liu, P. and Yang, Y. and Callan, J. and Neubig, G.},
  title     = {{PAL}: Program-Aided Language Models},
  booktitle = {Proceedings of the 40th International Conference on Machine Learning},
  series    = {Proceedings of Machine Learning Research},
  volume    = {202},
  pages     = {10764--10799},
  publisher = {PMLR},
  year      = {2023}
}

@inproceedings{razeghi2022impact,
  author    = {Razeghi, Y. and Logan, R. L. IV and Gardner, M. and Singh, S.},
  title     = {Impact of Pretraining Term Frequencies on Few-Shot Numerical Reasoning},
  booktitle = {Findings of the Association for Computational Linguistics: EMNLP 2022},
  pages     = {840--854},
  publisher = {Association for Computational Linguistics},
  year      = {2022}
}

@article{garcez2019neuralsymbolic,
  author  = {Garcez, A. d. and Gori, M. and Lamb, L. C. and Serafini, L. and Spranger, M. and Tran, S. N.},
  title   = {Neural-Symbolic Computing: An Effective Methodology for Principled Integration of Machine Learning and Reasoning},
  journal = {Journal of Applied Logics},
  volume  = {6},
  number  = {4},
  pages   = {611--632},
  year    = {2019}
}

@article{kautz2022third,
  author  = {Kautz, H.},
  title   = {The Third AI Summer: AAAI Robert S. Engelmore Memorial Lecture},
  journal = {AI Magazine},
  volume  = {43},
  number  = {1},
  pages   = {105--125},
  year    = {2022}
}

@inproceedings{lyu2023faithfulcot,
  author    = {Lyu, Q. and Havaldar, S. and Stein, A. and Zhang, L. and Rao, D. and Wong, E. and Apidianaki, M. and Callison-Burch, C.},
  title     = {Faithful Chain-of-Thought Reasoning},
  booktitle = {Proceedings of the 13th International Joint Conference on Natural Language Processing and the 3rd Conference of the Asia-Pacific Chapter of the Association for Computational Linguistics (Volume 1: Long Papers)},
  pages     = {305--329},
  address   = {Nusa Dua, Bali},
  publisher = {Association for Computational Linguistics},
  month     = nov,
  year      = {2023},
  doi       = {10.18653/v1/2023.ijcnlp-main.20}
}

@inproceedings{vakharia2024proslm,
  author    = {Vakharia, P. and Kufeldt, A. and Meyers, M. and Lane, I. and Gilpin, L. H.},
  title     = {{ProSLM}: A Prolog Synergized Language Model for Explainable Domain Specific Knowledge Based Question Answering},
  booktitle = {Proceedings of the 2024 International Conference on Neural-Symbolic Learning and Reasoning (NeSy)},
  pages     = {291--304},
  year      = {2024}
}

@article{mtl-overview,
  author  = {Caruana, R.},
  title   = {Multitask Learning},
  journal = {Machine Learning},
  volume  = {28},
  number  = {1},
  pages   = {41--75},
  year    = {1997},
  doi     = {10.1023/A:1007379606734}
}

@inproceedings{data-mixing,
  author    = {Li, Y. and Liu, Z. and Xing, E.},
  title     = {Data Mixing Optimization for Supervised Fine-Tuning of Large Language Models},
  booktitle = {Proceedings of the 42nd International Conference on Machine Learning},
  series    = {Proceedings of Machine Learning Research},
  volume    = {267},
  pages     = {35419--35437},
  publisher = {PMLR},
  year      = {2025}
}

@inproceedings{wei2022finetuned,
  author    = {Wei, J. and Bosma, M. and Zhao, V. Y. and Guu, K. and Yu, A. W. and Lester, B. and Du, N. and Dai, A. M. and Le, Q. V.},
  title     = {Finetuned Language Models Are Zero-Shot Learners},
  booktitle = {International Conference on Learning Representations},
  year      = {2022}
}

@inproceedings{sanh2022multitask,
  author    = {Sanh, V. and Webson, A. and Raffel, C. and others},
  title     = {Multitask Prompted Training Enables Zero-Shot Task Generalization},
  booktitle = {International Conference on Learning Representations},
  year      = {2022}
}

@inproceedings{wang2023codet5plus,
  author    = {Wang, Y. and Le, H. and Gotmare, A. D. and Bui, N. D. Q. and Li, J. and Hoi, S. C. H.},
  title     = {{CodeT5+}: Open Code Large Language Models for Code Understanding and Generation},
  booktitle = {Proceedings of the 2023 Conference on Empirical Methods in Natural Language Processing},
  pages     = {1069--1088},
  publisher = {Association for Computational Linguistics},
  year      = {2023},
  doi       = {10.18653/v1/2023.emnlp-main.68}
}

@article{chen2021codex,
  author  = {Chen, M. and Tworek, J. and Jun, H. and others},
  title   = {Evaluating Large Language Models Trained on Code},
  journal = {CoRR},
  volume  = {abs/2107.03374},
  year    = {2021}
}

@article{li2022alphacode,
  author  = {Li, Y. and Choi, D. and Chung, J. and Kushman, N. and Schrittwieser, J. and Leblond, R. and Eccles, T. and Keeling, J. and Gimeno, F. and Dal Lago, A. and others},
  title   = {Competition-Level Code Generation with AlphaCode},
  journal = {Science},
  volume  = {378},
  number  = {6624},
  pages   = {1092--1097},
  year    = {2022},
  doi     = {10.1126/science.abq1158}
}

@article{codellama,
  author  = {Rozi{\`e}re, B. and others},
  title   = {Code Llama: Open Foundation Models for Code},
  journal = {arXiv preprint arXiv:2308.12950},
  year    = {2023}
}

@inproceedings{chen2023teaching,
  author    = {Chen, X. and Lin, M. and Sch{\"a}rli, N. and Zhou, D.},
  title     = {Teaching Large Language Models to Self-Debug},
  booktitle = {International Conference on Learning Representations},
  year      = {2024}
}

@inproceedings{shinn2023reflexion,
  author    = {Shinn, N. and Cassano, F. and Berman, E. and Gopinath, A. and Narasimhan, K. and Yao, S.},
  title     = {Reflexion: Language Agents with Verbal Reinforcement Learning},
  booktitle = {Advances in Neural Information Processing Systems},
  volume    = {36},
  year      = {2023}
}

@inproceedings{yang2024arithmetic,
  author    = {Yang, X. and Chen, B. and Tam, Y.-C.},
  title     = {Arithmetic Reasoning with LLM: Prolog Generation \& Permutation},
  booktitle = {Proceedings of the 2024 Conference of the North American Chapter of the Association for Computational Linguistics: Human Language Technologies (Volume 2: Short Papers)},
  pages     = {699--710},
  address   = {Mexico City, Mexico},
  publisher = {Association for Computational Linguistics},
  year      = {2024},
  doi       = {10.18653/v1/2024.naacl-short.61}
}

@misc{brief2024mixing,
  author        = {Brief, M. and Ovadia, O. and Shenderovitz, G. and Ben Yoash, N. and Lemberg, R. and Sheetrit, E.},
  title         = {Mixing It Up: The Cocktail Effect of Multi-Task Fine-Tuning on LLM Performance -- A Case Study in Finance},
  year          = {2024},
  publisher     = {arXiv},
  eprint        = {2410.01109},
  archivePrefix = {arXiv},
  primaryClass  = {cs.AI},
  doi           = {10.48550/arXiv.2410.01109},
  url           = {https://arxiv.org/abs/2410.01109}
}


\end{document}